\documentclass[runningheads]{llncs}
\usepackage{amsmath}
\usepackage[colorlinks=true, allcolors=blue]{hyperref}
\usepackage{bm}
\usepackage{stmaryrd}
\usepackage{graphicx}
\begin{document}
\title{A unified view on Self-Organizing Maps (SOMs) \\and Stochastic Neighbor Embedding (SNE)}
\titlerunning{A unified view on SOMs and SNE}
%
\author{Thibaut Kulak \and Anthony Fillion \and François Blayo}
\authorrunning{T. Kulak et al.}
%
\institute{NeoInstinct S.A.\\ Rue traversière 3, 1018 Lausanne (Switzerland)\\
\email{thibaut.kulak@neoinstinct.com}}
\maketitle              
\begin{abstract}
We propose a unified view on two widely used data visualization techniques: Self-Organizing Maps (SOMs) and Stochastic Neighbor Embedding (SNE). We show that they can both be derived from a common mathematical framework. Leveraging this formulation, we propose to compare SOM and SNE quantitatively on two datasets, and discuss possible avenues for future work to take advantage of both approaches.

\end{abstract}
\section{Introduction}
Self-organizing Maps (SOMs) \cite{kohonen1990self} have been widely used over the last decades on diverse applications \cite{van2012self}, and have emerged as a very efficient biologically-inspired method for learning organized representations of data. More recently, the Stochastic Neighbor Embedding (SNE) algorithm  \cite{hinton2002stochastic} has been proposed to visualize a set of datapoints in a two-dimensional space, hence creating a map. This algorithm, and notably its extension t-distributed Stochastic Neighbor Embedding (t-SNE) \cite{van2008visualizing} has become very popular for visualizing large datasets. The goal of this work is to propose a unified view over those two algorithms.\\

We will show that both SOM and SNE algorithms can be seen in the context of a general mathematical formulation that attempts to find points in a 2D space associated to weights in an observation space, where the weights represent the data, and where neighborhood relations in the 2D space and the observation space are respected (points that are close in the map should be close in the observation space, and vice versa). In the context of this optimization problem, we will see that SOM's algorithm fixes the points in the 2D space, and then optimizes the weights of the neurons, while SNE fixes the weights of the neurons, and optimizes the positions of the points in the 2D space.\\

Such a unified view of those algorithms brings several contributions: for SOMs, this sheds light on what is actually minimized and on how we might be able to assess the quality of a map both in terms of organization and representation, which remains so far non-trivial \cite{de2002statistical,bauer1992quantifying}. For SNE, we propose a formulation that showcases the fact that SNE is not only a visualization technique but is also learning a representation of the data. Moreover, the number of neuron weights can be different from the number of datapoints in our formulation, which goes beyond the standard SNE approach that assumes one neuron per datapoint. Most importantly, as we will discuss later, we hope that bridging those algorithms will open the way to future works leveraging their respective advantages.\\

Our paper is organized as follows: first, we introduce the mathematical framework, and show that the SOM and SNE algorithms can be derived from it. Then, we exploit the fact that both algorithms pursue a very similar objective to compare their results quantitatively on two different datasets. Finally, we conclude and discuss interesting avenues for future work leveraging our unified formulation.

\subsection{Notations and definitions}

We assume there are $N$ points $\{z_{i}\}_{i=1}^N$ in a 2D space (visualization space), corresponding to $N$ weights (neuron weights) $\{w_{i}\}_{i=1}^N$ in a $d$-dimensional space (observation space). We introduce two conditional distributions $p(i|x)$ and $q(i|x)$ defined for a datapoint (stimulus) $x$. Those distributions are defined over the discrete set $\llbracket 1;N \rrbracket$:

\begin{equation}
    p(i|x) = \frac{\exp{\Big(-\frac{||x-w_{i}||^2}{\sigma^2}}\Big)}{\sum_{j=1}^N \exp{\Big(-\frac{||x-w_{j}||^2}{\sigma^2}} \Big) } \quad \text{and} \quad
    q(i|x) = \frac{\exp{ \Big( -||z_{i}-z_{i^{*}(x)}||^2} \Big)}{\sum_{j=1}^N \exp{\Big( -||z_{j}-z_{i^{*}(x)}||^2 \Big)}},
\end{equation}
where $\sigma$ is an hyperparameter\footnote{Note that $p(i|x)$ has been obtained from Bayes rule by choosing $p(x|i) = \mathcal{N}(w_{i}, \sigma^2)$ and a uniform prior over the discrete space $p(i) = 1/N$, so $\sigma$ is actually the chosen standard deviation of $p(x|i)$. Such model also defines a probability distribution on the observation space $p(x) = \sum_{i=1}^N p(x|i) p(i) = 1/N \sum_{i=1}^N p(x|i)$}, and $i^{*}(x) = \arg \max_{i=1}^N p(i|x)$ is the index of the winning neuron (the neuron that responds the most to the stimulus $x$).

We can interpret those conditional distributions qualitatively:
\begin{itemize}
    \item $p(i|x)$ increases with the proximity of the weights to the stimulus \textbf{in the observation space}, i.e., which neurons respond to the stimulus.
    \item $q(i|x)$ increases with the proximity of the points to the stimulus \textbf{in the 2D (visualization) space}
\end{itemize}

\begin{figure}
\begin{minipage}{0.48\textwidth}
\begin{center}
\includegraphics[width=0.9\textwidth]{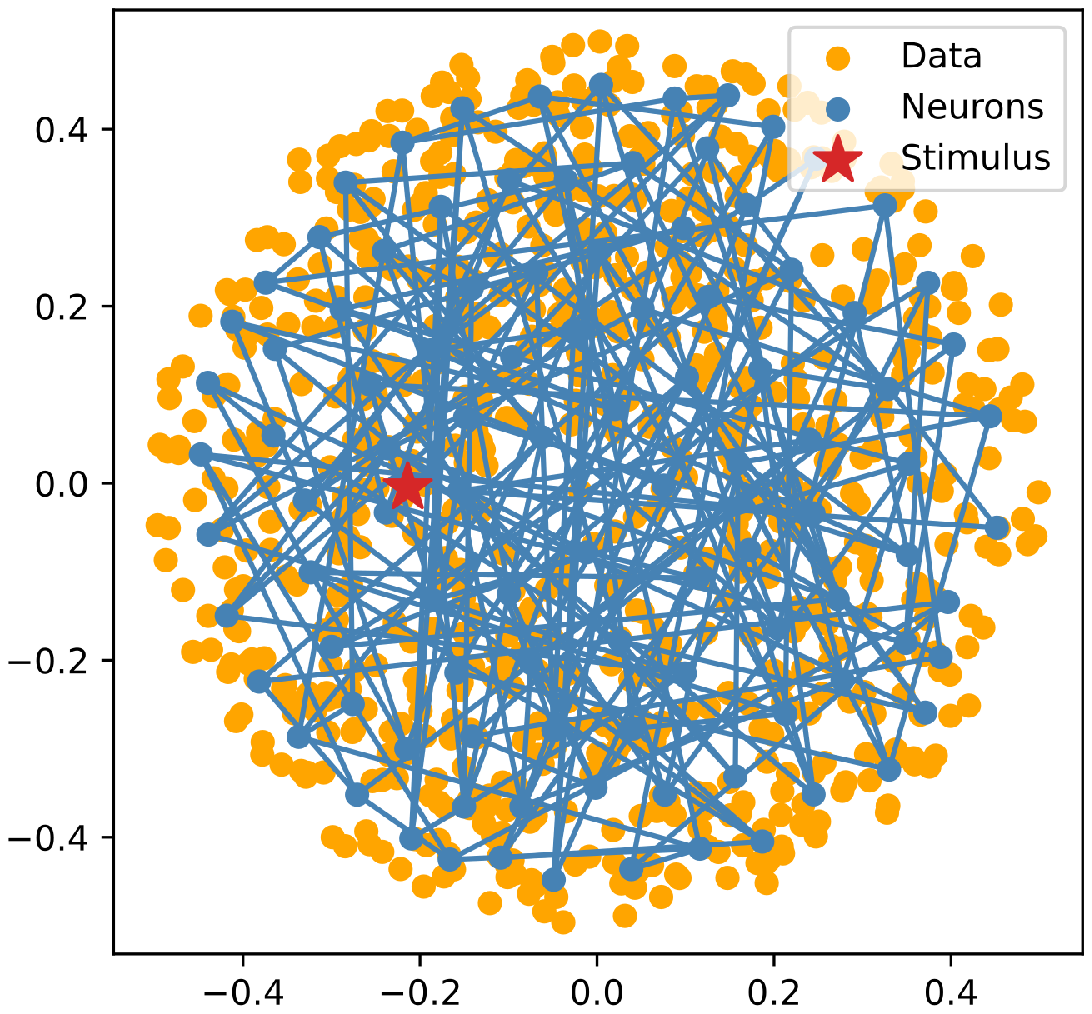}\\
\includegraphics[width=0.99\textwidth]{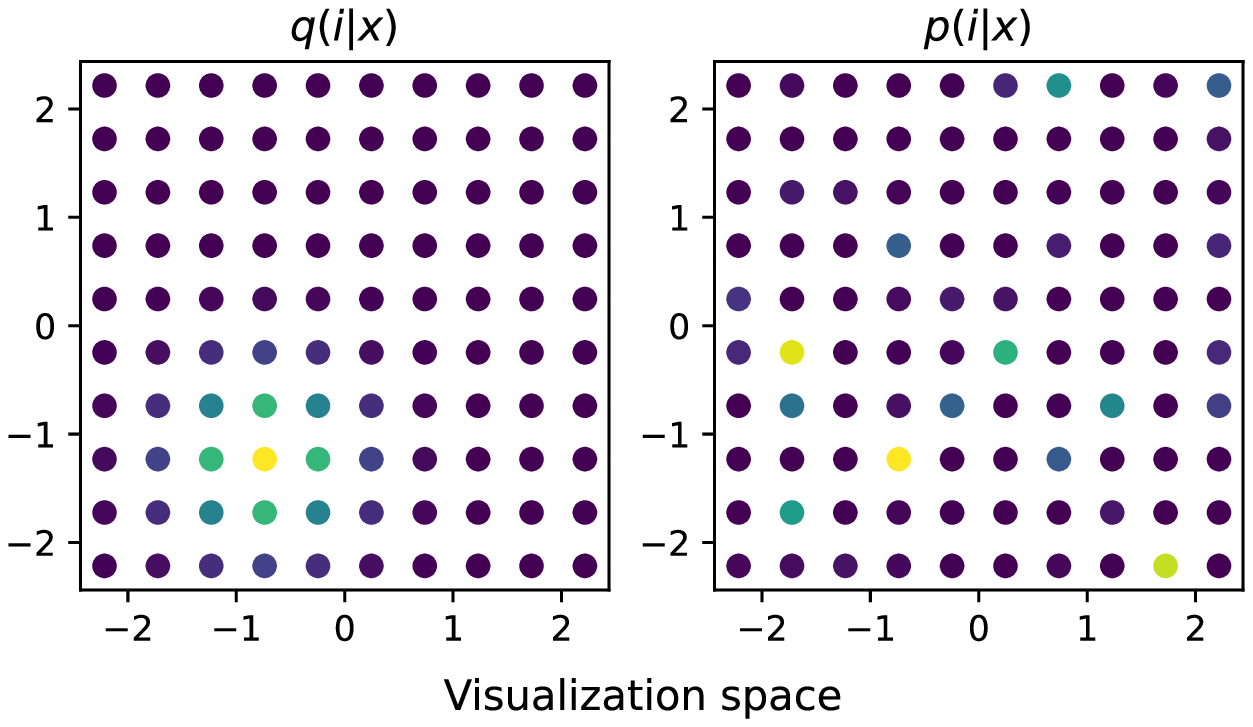}\\
\includegraphics[width=\textwidth]{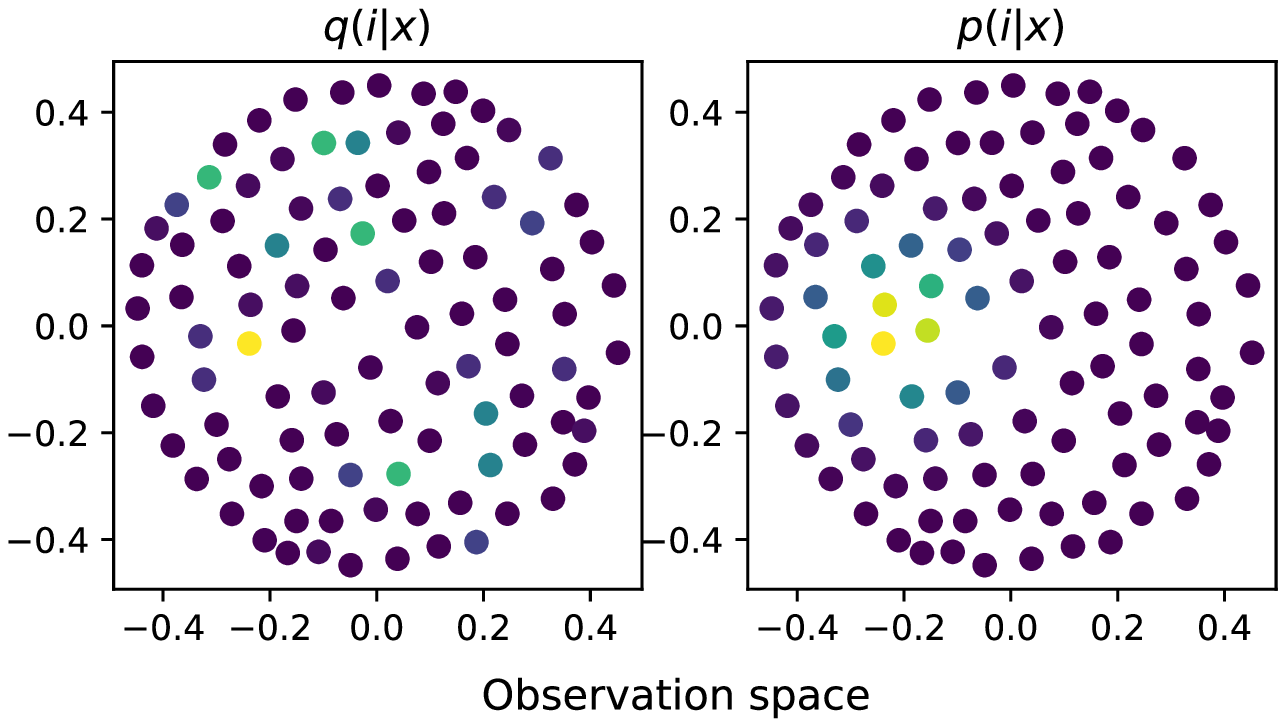}
\end{center}
\end{minipage}
\hspace{\stretch{1}}
\begin{minipage}{0.48\textwidth}
\begin{center}
\includegraphics[width=0.9\textwidth]{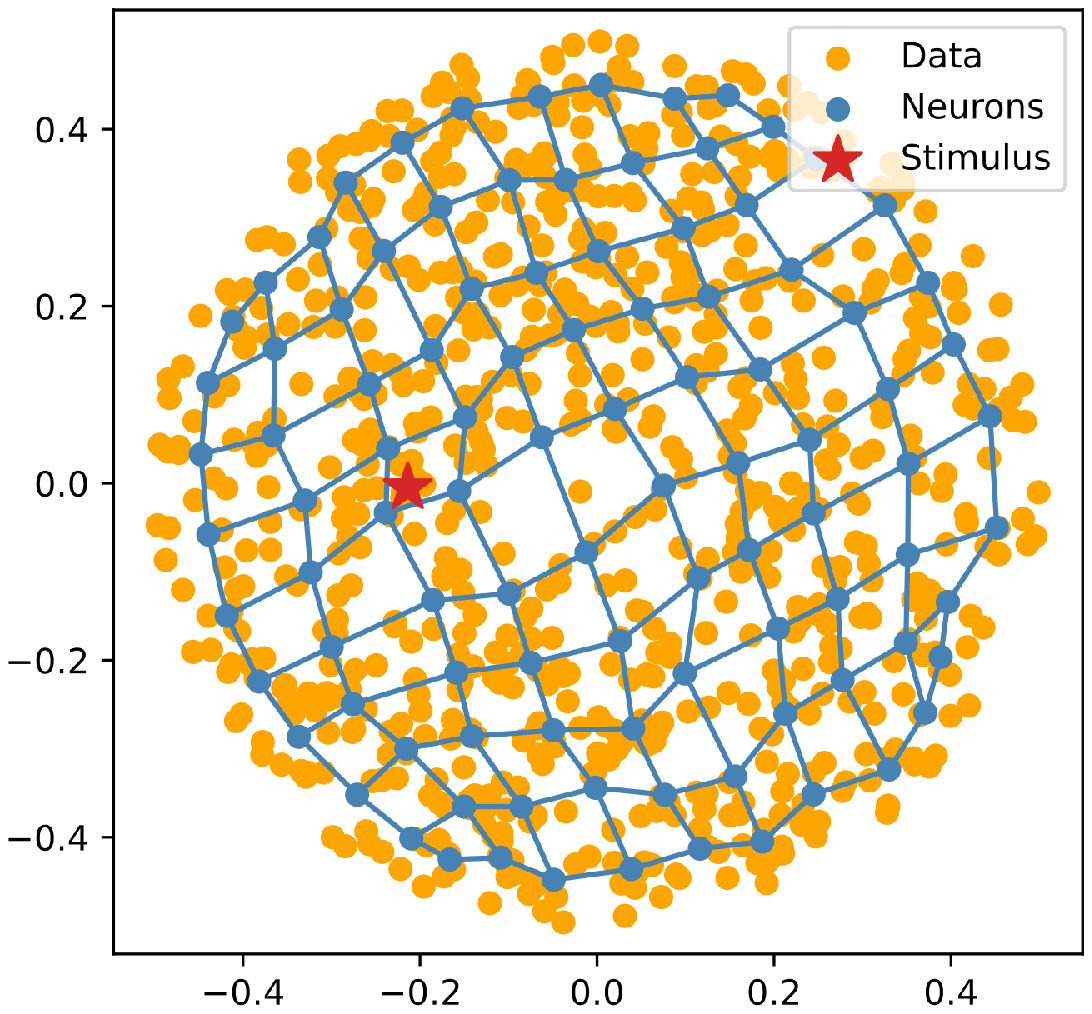}\\
\includegraphics[width=0.99\textwidth]{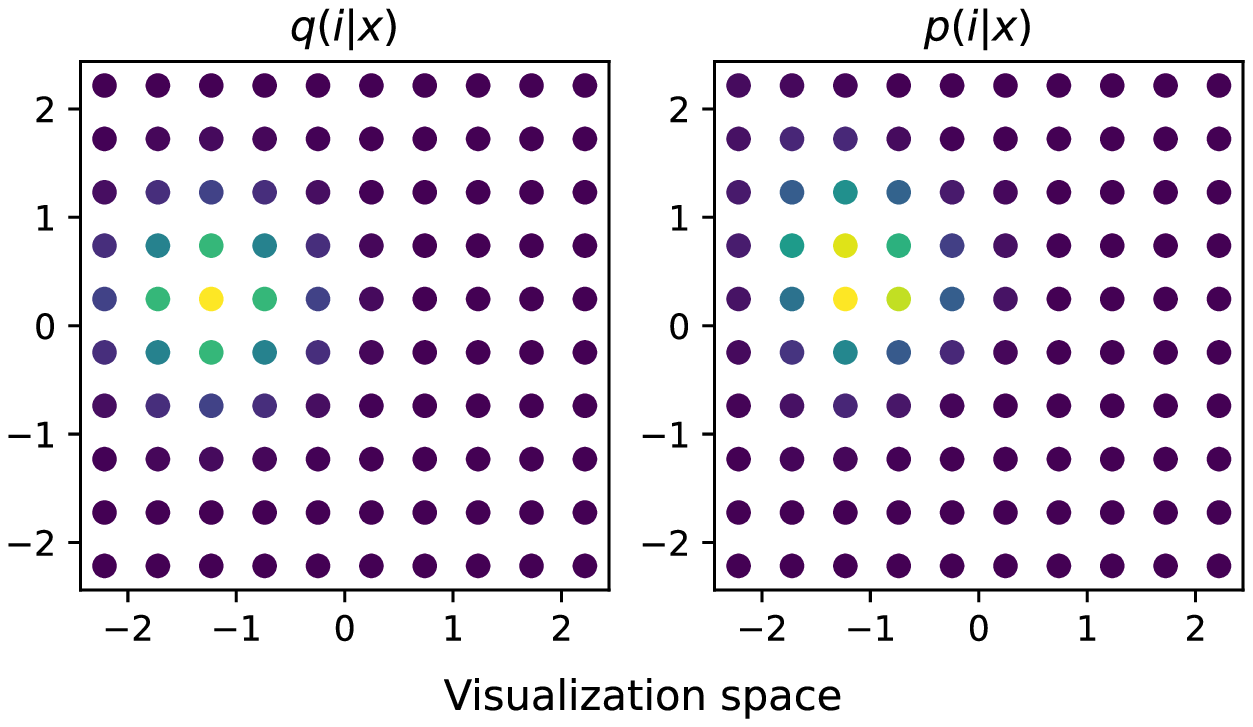}\\
\includegraphics[width=\textwidth]{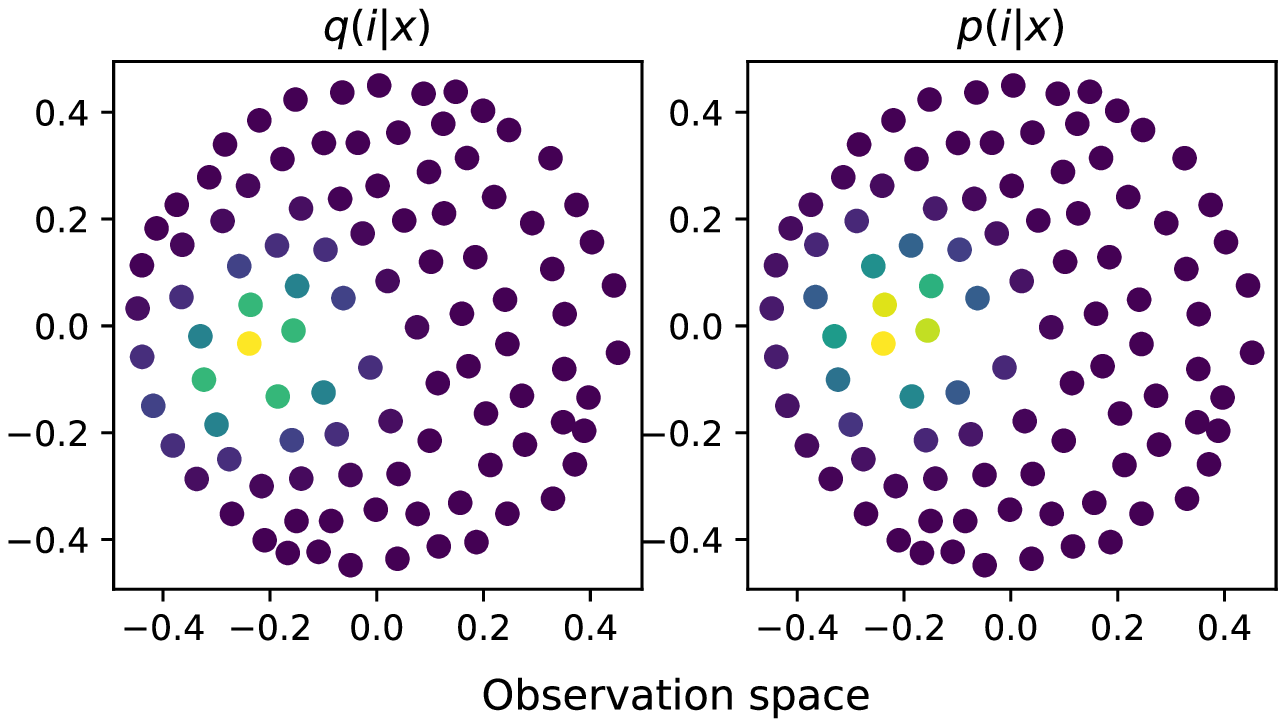}
\end{center}
\end{minipage}
\caption{Visualization of an unorganized map (left) and an organized map (right), along with the conditional distributions $p(i|x)$ and $q(i|x)$, both in visualization space (points $\{z_{i}\}$ colored by the probabilities, middle row) and observation space (weights $\{w_{i}\}$ colored by the probabilities, bottom row). Purple color corresponds to low probabilities and yellow to high probabilities}
\label{fig:conditionaldistributions}
\end{figure}

To give further intuition on those conditional distributions, we show in Fig.~\ref{fig:conditionaldistributions} those distributions for two different maps, one organized map that was obtained using SOM's algorithm and one non-organized map (that was obtained by randomly permuting the indices of the neurons weights of the organized map). We can see that when the map is organized $p(i|x)$ and $q(i|x)$ are close to each other, which is not the case for the non-organized map. More precisely, if we look at $p(i|x)$ and $q(i|x)$ in the visualization space, $q(i|x)$ looks like a Gaussian distribution no matter if the map is organized or not, but $p(i|x)$ looks like a Gaussian only if the map is organized, which indeed means that points around the winning neuron also respond to the stimulus. Similarly, if we look at $p(i|x)$ and $q(i|x)$ in the observation space, $p(i|x)$ looks like a Gaussian distribution no matter if the map is organized or not, but $q(i|x)$ looks like a Gaussian only if the map is organized, which indeed means that the points that respond to the stimulus are those that are close in the visualization space. We will see in the next sections that both SOMs and SNE methods attempt to make these distributions close to each other, while representing the data.

\subsection{SOMs in this framework}

Let us assume that stimulus $x$'s are samples from an unknown density $p^{\text{data}}(x)$. We propose the following objective function, that is a trade-off between organization (in the sense of making $p(i|x)$ and $q(i|x)$ close) and representation (maximizing $p(x)$):

\begin{equation}
    J^{\text{SOM}}\Big(\{z_{i}\}_{i=1}^N, \{w_{i}\}_{i=1}^N\Big) = \int_{x} \Big( \mathcal{D}_{KL}(q(i|x), p(i|x)) - \ln p(x) \Big) \; p^{\text{data}}(x) \; dx
\end{equation}

The standard SOM algorithm (non-batch version) is a stochastic algorithm that updates the neuron weights as a stimulus $x$ is presented. For a stimulus $x$ coming from $p^{\text{data}}(x)$, we can get the objective function for this stimulus $x$ by removing the integral over the unknown density function $p^{\text{data}}(x)$ (stochastic gradient descent):

\begin{equation}
    J^{\text{SOM}}\Big(\{z_{i}\}_{i=1}^N, \{w_{i}\}_{i=1}^N \; | \; x \Big) = \mathcal{D}_{KL}(q(i|x), p(i|x)) - \ln p(x)
\label{eq:SOM_objective}
\end{equation}

Differentiating with respect to the neuron weights and negating gives us the following update rule of the neuron weights:

\begin{equation}
    - \frac{\partial J^{\text{SOM}}\Big(\{z_{i}\}_{i=1}^N, \{w_{i}\}_{i=1}^N \; | \; x \Big)}{\partial w_{i}} = \frac{2}{\sigma^2}(x - w_{i}) q(i|x)
\end{equation}

We can see that this is exactly the update rule of SOM's algorithm, where neuron weights get moved towards the stimulus $x$ if they are in the neighborhood of the winning neuron.\\

In SOM's algorithm, there is typically a neighborhood parameter that is monotically decreasing. In our framework we did not put a neighborhood parameter on $q(i|x)$, because we can alternatively see it\footnote{Trivially: $\frac{||z_{i}-z_{i^{*}(x)}||^2}{\sigma_{z}^2} = ||\frac{z_{i}}{\sigma_{z}}-\frac{z_{i^{*}(x)}}{\sigma_{z}}||^2$} as a scaling of the points $\{z_{i}\}_{i=1}^N$ typically chosen on a grid. Choosing a monotically decreasing neighborhood parameter is equivalent to scaling monotically the points $\{z_{i}\}_{i=1}^N$ from a small to a big grid with fixed neighborhood parameter.\\

In our view, the SOM algorithm is performing an optimization over a set of points in a 2D space and their corresponding weights, and is supposing a particular time-dependent form for the points locations in the 2D space (from a small to a big grid). Viewing it this way will make it easier to compare with SNE in the next section, which does not use any neighborhood parameter in the visualization space.

\subsection{SNE in this framework}

In this section, we will show that the SNE algorithm can be derived from our framework, hence show that both SOMs and SNE algorithms are solving the same mathematical problem, and exhibit their differences.\\

Let us define for a stimulus $x$ the following objective function:
\begin{equation}
    J^{\text{SNE}}\Big(\{z_{i}\}_{i=1}^N, \{w_{i}\}_{i=1}^N \; | \; x \Big) = \mathcal{D}_{KL}(p(i|x), q(i|x)) - \ln p(x)
\label{eq:SNE_objective}
\end{equation}

This objective is very similar to Eq.~\ref{eq:SOM_objective}, the only difference is the use of the forward KL divergence instead of the reverse KL divergence. Given that the KL divergence is not symmetric, Eq.~\ref{eq:SOM_objective} and Eq.~\ref{eq:SNE_objective} differ only by the way they measure the difference between $p(i|x)$ and $q(i|x)$.\\

Differentiating with respect to the 2D points locations gives us the following update rule of the points:

\begin{equation}
- \frac{\partial J^{\text{SNE}}\Big(\{z_{i}\}_{i=1}^N, \{w_{i}\}_{i=1}^N \; | \; x \Big)}{\partial z_{i}} = 2(z_{i^{*}(x)} - z_{i}) (q(i|x) - p(i|x))
\end{equation}

This update rule is very similar from the SNE algorithm and can be interpreted as such: if a neuron weight $w_{i}$ is far from the stimulus $x$ (i.e., $p(i|x)$ small), but is close to the winning neuron in the 2D visualization space (i.e., $q(i|x)$ big) then the point is moved towards the location of the winning neuron in the 2D space. Reversely, if a neuron weight $w_{i}$ is close from the stimulus $x$ (i.e., $p(i|x)$ big), but is far from the winning neuron in the 2D visualization space (i.e., $q(i|x)$ small) then the point is moved away from the location of the winning neuron in the 2D space.\\

We have presented here the update rule equations for any set of neuron weights. In the standard SNE algorithm, usually the neuron weights are equal to the datapoints, because the number of neurons chosen is equal to the number of datapoints to visualize. In that case, a stimulus $x$ is therefore a weight $w_{j}$, and trivially $i^{*}(x) = j$, which gives us the usual SNE formulation.\footnote{In practice, the SNE formulation is slightly different because it is done in batch mode (using all datapoints $x$ for each update), the conditional distributions $p(i|w_{j})$ and $q(i|w_{j})$ are set to zero for $i=j$, and a different $\sigma_{i}$ is used for each stimulus $x=w_{i}$ to account for different local variations in the data.}

It is important to note that Eq.~\ref{eq:SNE_objective} differs from the standard SNE objective function by the addition of a representation term $-\ln p(x)$, which only depends on the neuron weights $\{w_{i}\}_{i=1}^N$ and not on the 2D points, which is why we recover the same update rule as SNE. If, similarly to our interpretation of SOM's algorithm, we interpret SNE as an optimization over both the neuron weights and the 2D points, this term makes a lot of sense because the first step in SNE consists in choosing as many neurons as the number of datapoints and putting a Gaussian $p(x|i)$ around it. This can be seen as a kernel density estimation method, and hence as a minimization of $-\ln p(x)$\\

We have shown that SOMs and SNE are two dual algorithms sharing the same global objective function over the neuron weights and 2D points, the main difference being in the way they approach this optimization. SOMs fix the 2D points (typically, on a grid) and optimize the neuron weights, while SNE fixes the neuron weights (typically one for datapoint) and optimizes the 2D points locations. We outline their differences in Table~\ref{table:comparison}.

{\renewcommand{\arraystretch}{1.75}
\begin{table}[]
\caption{Main differences between SOM and SNE}
    \centering
    \begin{tabular}{|c|c|c|}
    \hline
     & \textbf{SOM} & \textbf{SNE} \\
     \hline
     Objective & Reverse KL + Data fitting &  Forward KL + Data fitting \\[-5pt]
     & $\mathcal{D}_{KL}(q(i|x), p(i|x)) - \ln p(x)$ & $\mathcal{D}_{KL}(p(i|x), q(i|x)) - \ln p(x)$ \\
     \hline
     Given & 2D points $\{z_{i}\}_{i=1}^N$ are given & Neuron weights $\{w_{i}\}_{i=1}^N$ are given \\[-5pt]
      & (typically on a grid) & (typically one for each datapoint) \\
     \hline
     $\quad$Optimized$\quad$ & Neuron weights $\{w_{i}\}_{i=1}^N$ are learned & 2D points $\{z_{i}\}_{i=1}^N$ are learned \\[-5pt]
      & (with stochastic gradient descent) & (with gradient descent)\\
     \hline
    \end{tabular}
    \label{table:comparison}
\end{table}}

\section{Quantitative comparisons}
We have shown previously that the SOM and SNE algorithms pursue a very similar objective: making the conditional distribution of the neurons response to a stimulus $p(i|x)$ and the neighborhood of the winning neuron $q(i|x)$ close to each other, and fitting the data. Such unified view permits to compare quantitatively the quality of maps learned with SOM and SNE, and shed light on their performances. This section is organized as follows: first, we detail in more details our approach for comparing those techniques. Then we compare them on a toy two-dimensional dataset. Finally, we compare them on a more realistic real-world dataset, the Fashion-MNIST dataset.

\subsection{Comparison approach}
The objectives $J^{\text{SOM}}$ and $J^{\text{SNE}}$ we have introduced in previous section depend on the hyperparameter $\sigma$ governing what it means to be close in the observation space, and on the scaling of the 2D points governing what it means to be close in the visualization space. We propose to take inspiration from \cite{villmann1997topology} and \cite{demartines1997curvilinear} to choose those parameters in a meaningful way. The entropies in bits of the conditional distributions $p(i|x)$ and $q(i|x)$ are continuous approximations of the number of neighbors of the stimulus $x$ (a.k.a. perplexities \cite{van2008visualizing}), respectively in the observation and the visualization spaces. The method proposed in \cite{villmann1997topology} for comparing the organization of maps was to look at $k$ neighbors in the observation space, and look if they are indeed the $k$ neighbors in the visualization space, and vice versa, for a varying number of neighbors $k$. We propose to take a similar approach and, for a stimulus $x$ and a given perplexity $\text{Perp}$ (continuous number of neighbors), choose $\sigma$ so that $2^{H(p(i|x)} = \text{Perp}$, and scale the points $\{z_{i}\}_{i=1}^N$ so that $2^{H(q(i|x)} = \text{Perp}$. We will monitor $J^{\text{SOM}}$ and $J^{\text{SNE}}$ for different perplexities averaged over the datasets , to assess the quality of the map at different scales. Such approach is therefore a continuous equivalent of the topographic function introduced in \cite{villmann1997topology} for assessing the quality of a map organization, but quantifies the data representation at the same time because the objectives we proposed contain not only an organization term but also a data fitting term.\\

In all experiments, we have used an exponential decay for the neighborhood parameter (inverse of the scaling parameter) and the learning rate\footnote{$\sigma(t) = \sigma_{i} \big(\frac{\sigma_{f}}{\sigma_{i}}\big)^{t/t_{\text{max}}}$ and $\lambda(t) = \lambda_{i} \big(\frac{\lambda_{f}}{\lambda_{i}}\big)^{t/t_{\text{max}}}$.} for the SOM algorithm as suggested in \cite{rougier2011dynamic}. For SNE, We used the open-source implementation of t-SNE \cite{opentsne} with a number of degrees of freedom equal to 10000 (hence, considering a Gaussian distribution instead of a t-distribution, as in SNE) and the other default hyperparameters.

\subsection{Toy 2D datasets}
\begin{figure}
\begin{center}
\begin{minipage}{0.48\textwidth}
\begin{center}
\includegraphics[width=\textwidth]{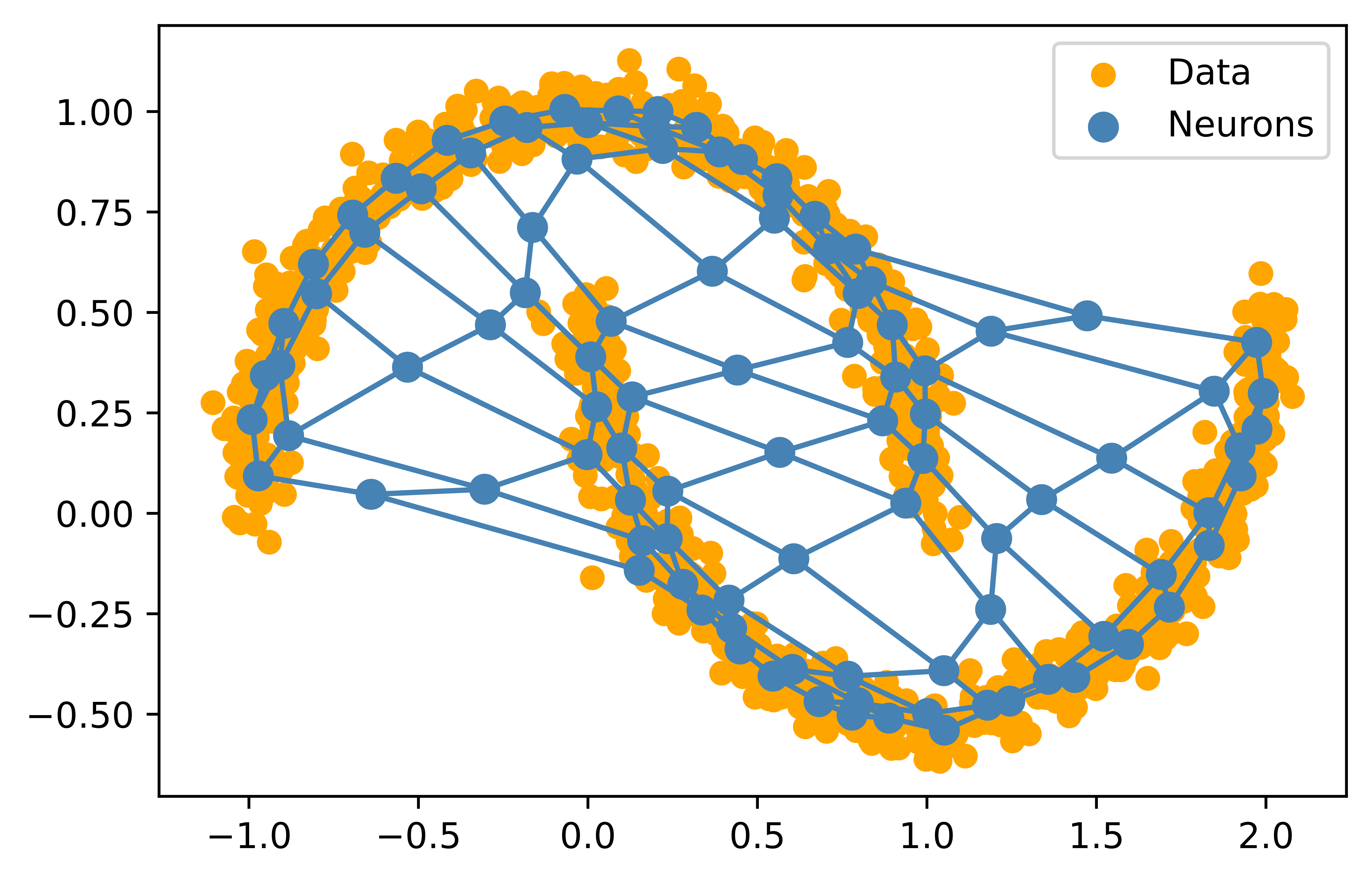}
SOM
\end{center}
\end{minipage}
\hspace{\stretch{1}}
\begin{minipage}{0.48\textwidth}
\begin{center}
\includegraphics[width=\textwidth]{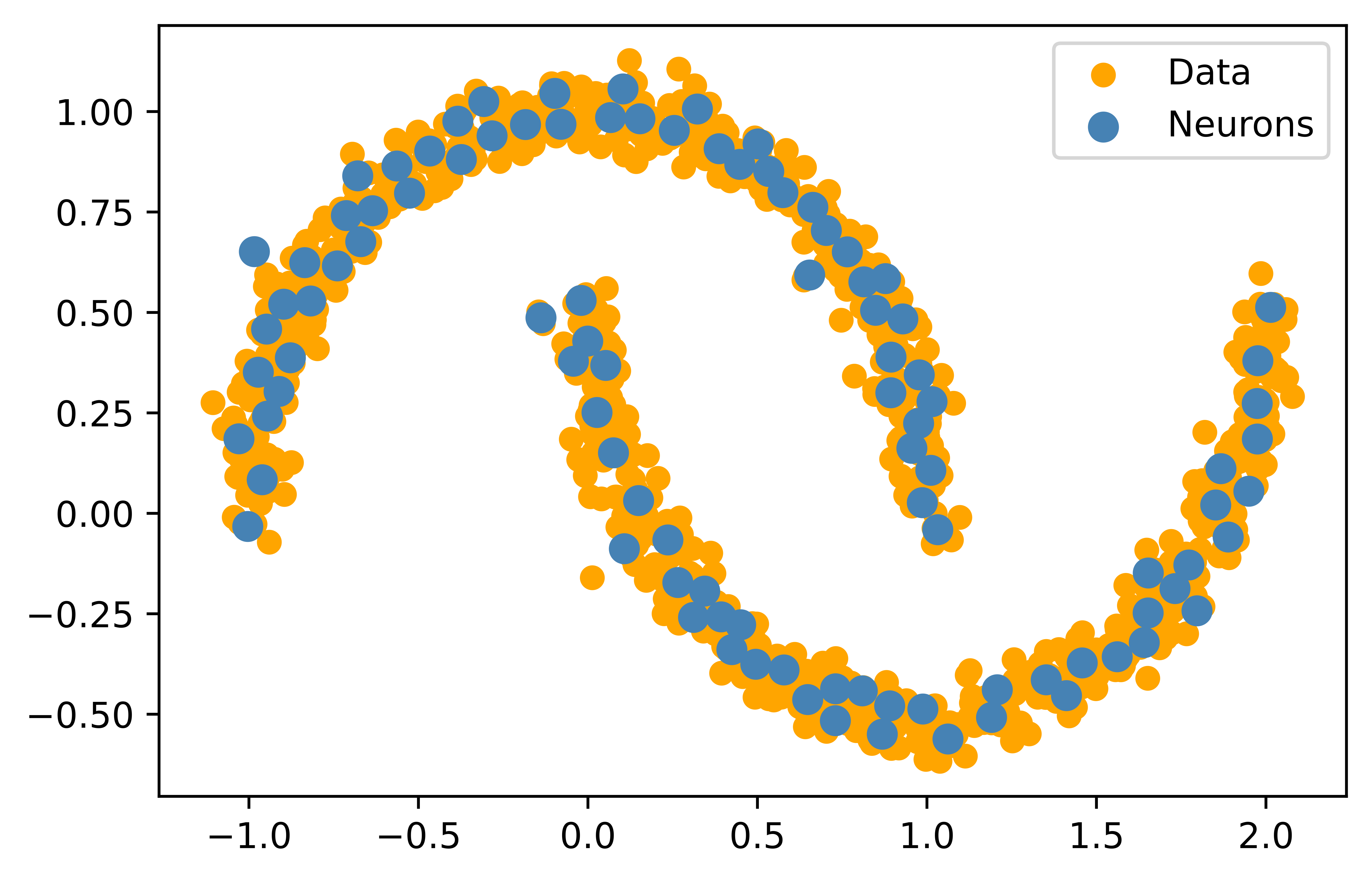}
kMeans+SNE
\end{center}
\end{minipage}

\vspace{0.2\baselineskip} Observation space \vspace{0.5\baselineskip}

\begin{minipage}{0.48\textwidth}
\begin{center}
\includegraphics[width=\textwidth]{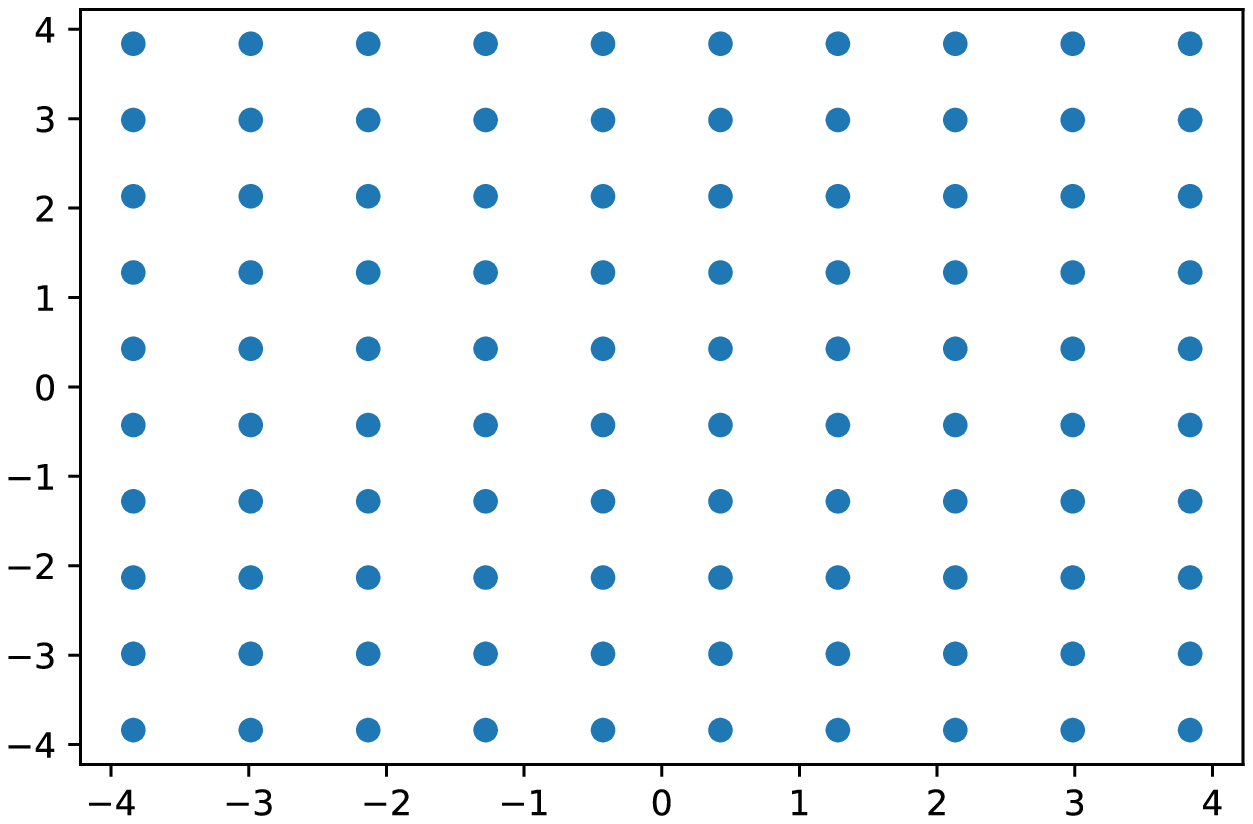}
SOM
\end{center}
\end{minipage}
\hspace{\stretch{1}}
\begin{minipage}{0.48\textwidth}
\begin{center}
\includegraphics[width=\textwidth]{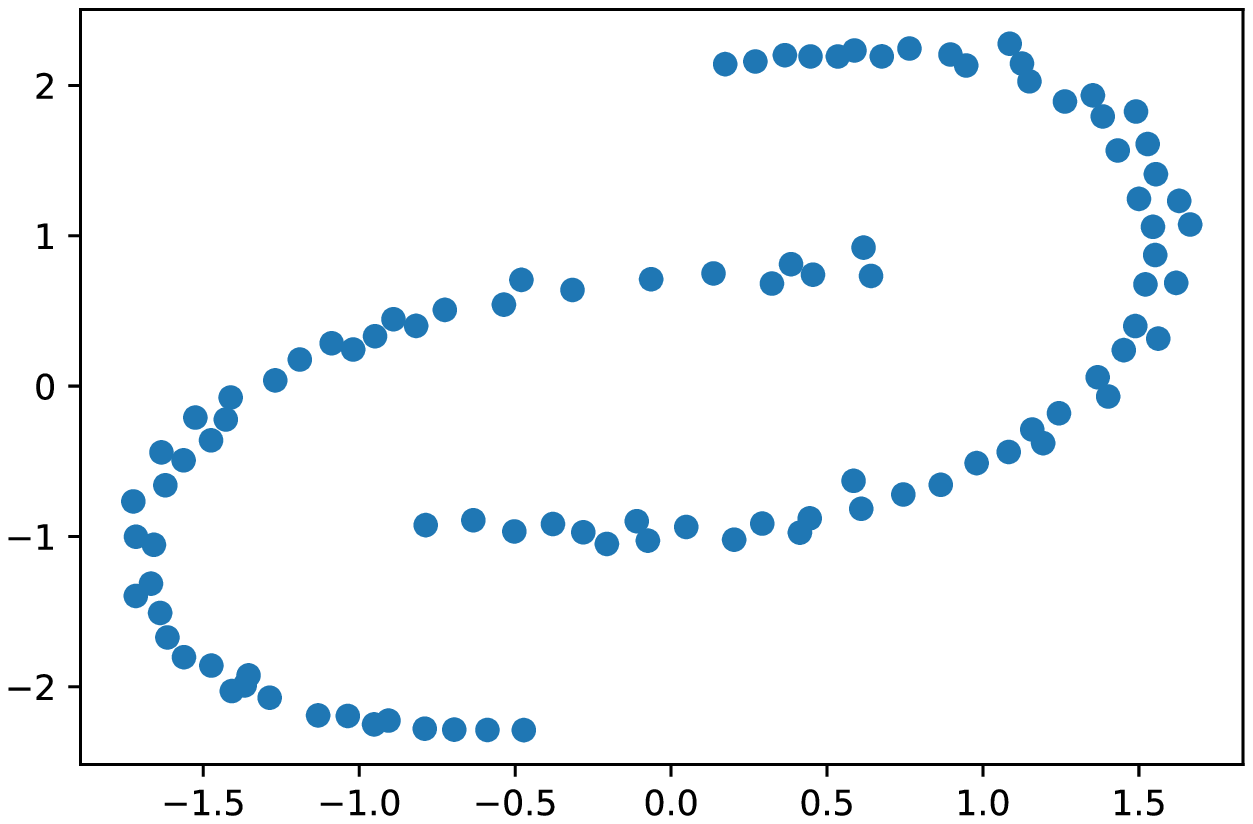}
kMeans+SNE
\end{center}
\end{minipage}
\vspace{0.2\baselineskip}\\ Visualization space
\end{center}
\caption{Maps learned on the \textbf{moons} dataset for the SOM and kMeans+SNE algorithms}
\label{fig:maps_moons}
\end{figure}

We compare here quantitatively the SOM and SNE algorithms on a toy two-dimensional dataset of 1000 points: the \textbf{moons} dataset. A visualization of the maps learned (in visualization and observation space) is shown in Fig.~\ref{fig:maps_moons}. A grid of $10\times10$ was used in these experiments\footnote{We chose $\sigma_{f}=10^{-10}$, $\sigma_{i}=100$, $\lambda_{f}=10^{-10}$, $\lambda_{i}=1$ and $t_{\text{max}}=100 000$}, for a fair comparison between SOM and SNE we did use also 100 neurons for SNE (and not one neuron per datapoint), those neuron weights have been learned using kMeans \cite{sklearn} with 100 clusters.

\begin{figure}
\begin{minipage}{0.48\textwidth}
\begin{center}
\includegraphics[width=\textwidth]{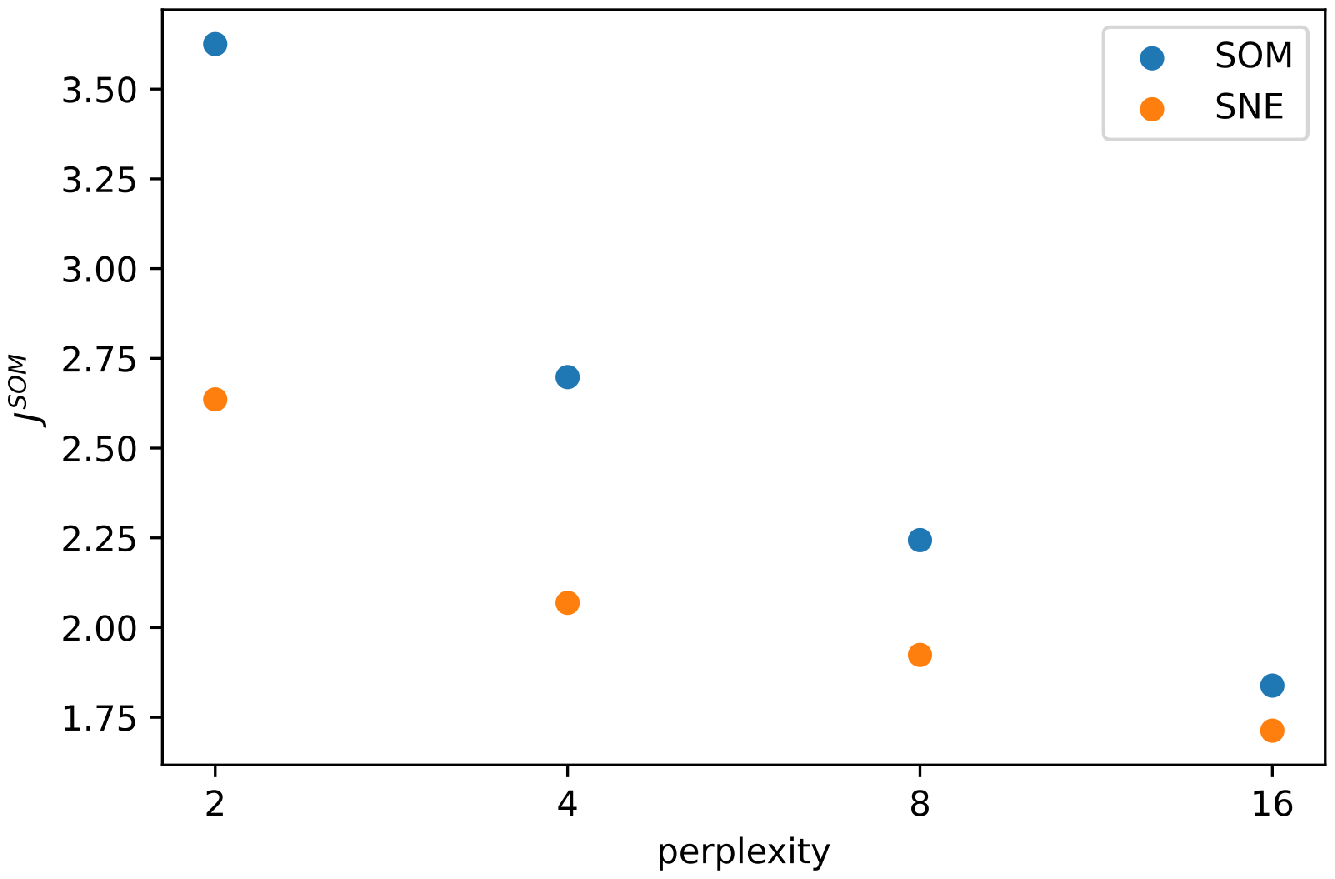}
\end{center}
\end{minipage}
\hspace{\stretch{1}}
\begin{minipage}{0.48\textwidth}
\begin{center}
\includegraphics[width=\textwidth]{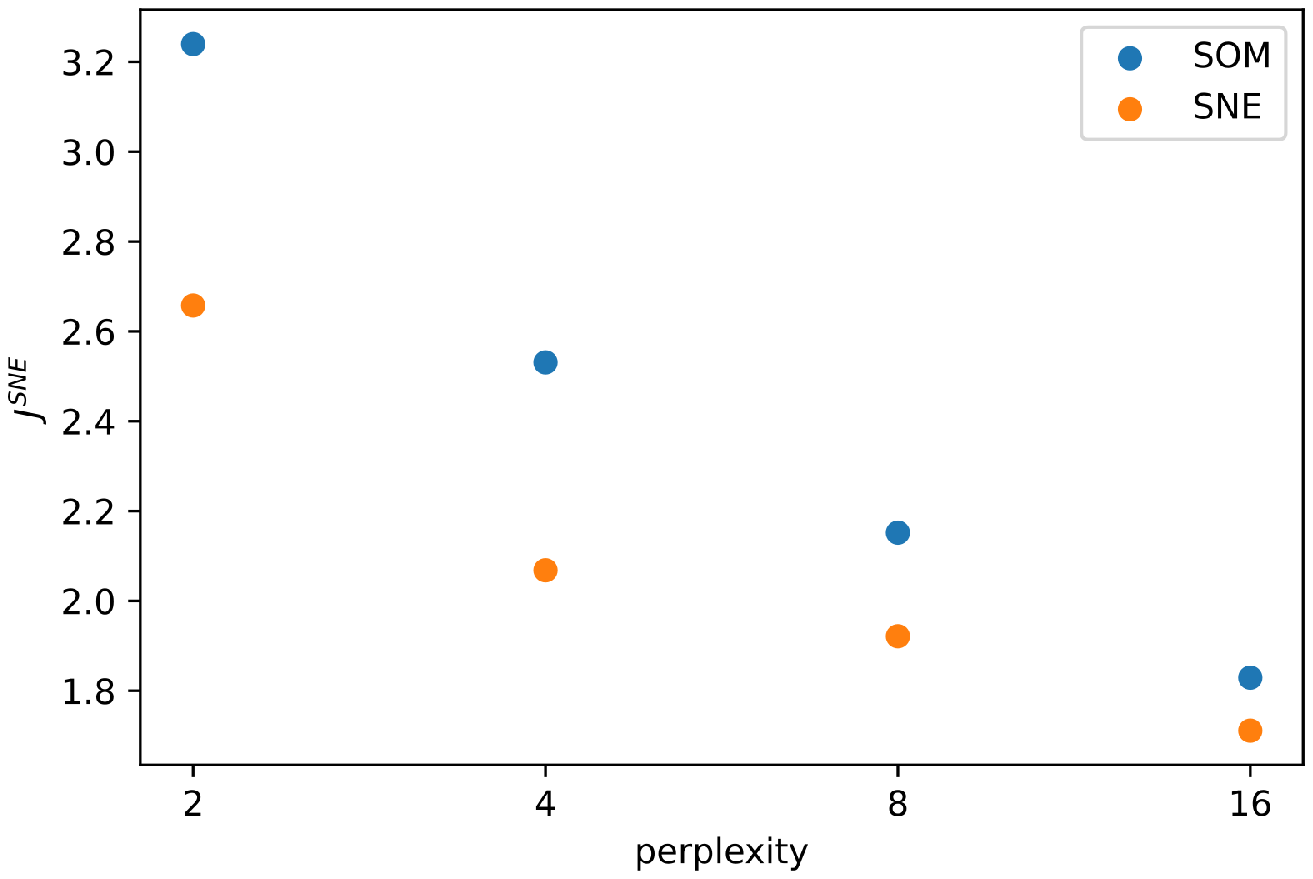}
\end{center}
\end{minipage}
\caption{Quantitative comparisons on the \textbf{moons} dataset of the maps learned with SOM and SNE, for the two objectives we proposed (left: SOM objective, right: SNE objective)}
\label{fig:results_moons}
\end{figure}

We can see in Fig.~\ref{fig:results_moons} that SNE clearly outperforms SOM for both objectives and all perplexities. We believe that this sheds light on the fact that assuming the points to lie on a grid is a very strong assumption of SOM's algorithm that drastically limits its performance capabilities in terms of the objectives defined above.

\subsection{Fashion-MNIST dataset}

We propose here to compare quantitatively the SOM and SNE algorithms on a more realistic dataset: a subset of 10k images taken randomly from the Fashion-MNIST dataset \cite{xiao2017}. We believe that SOM's assumption of the points lying on a grid might be less restrictive when using a very large number of neurons, which would be coherent with \cite{ultsch2005esom} arguing that large maps must be used for an interesting representation to emerge. We learn a $100\times100$ SOM\footnote{ We chose $\sigma_{f}=10^{-3}$, $\sigma_{i}=10$, $\lambda_{f}=10^{-3}$, $\lambda_{i}=1$ and $t_{\text{max}}=120 000$} on the Fashion-MNIST dataset, and as we chose 10k images from the original dataset we can use standard SNE with 10k neurons (one for each datapoint), and hence propose a clear and fair comparison between standard SOM and standard SNE with the same number of neurons.

\begin{figure}
\begin{minipage}{0.48\textwidth}
\begin{center}
\includegraphics[width=\textwidth]{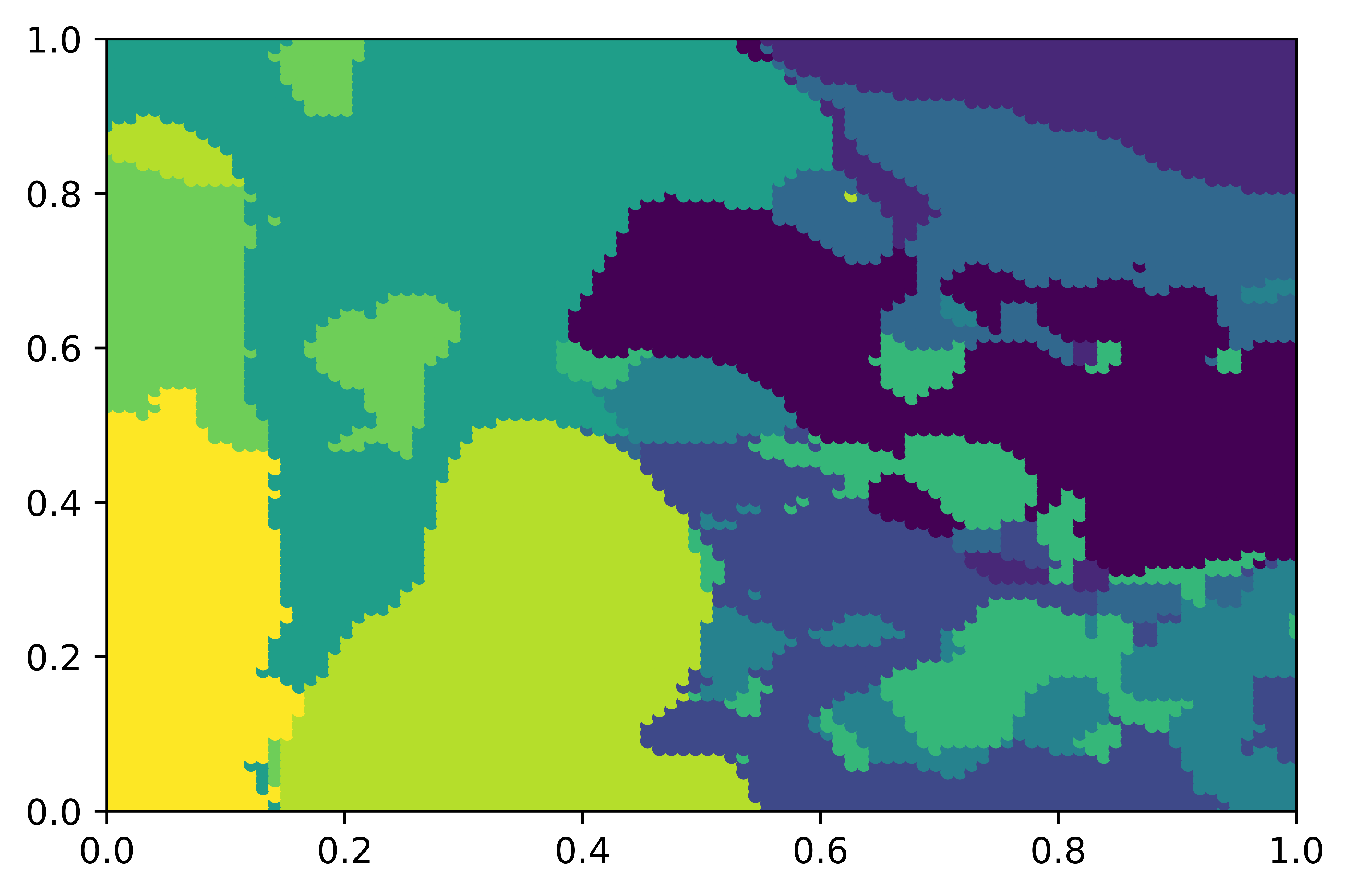}
SOM
\end{center}
\end{minipage}
\hspace{\stretch{1}}
\begin{minipage}{0.48\textwidth}
\begin{center}
\includegraphics[width=\textwidth]{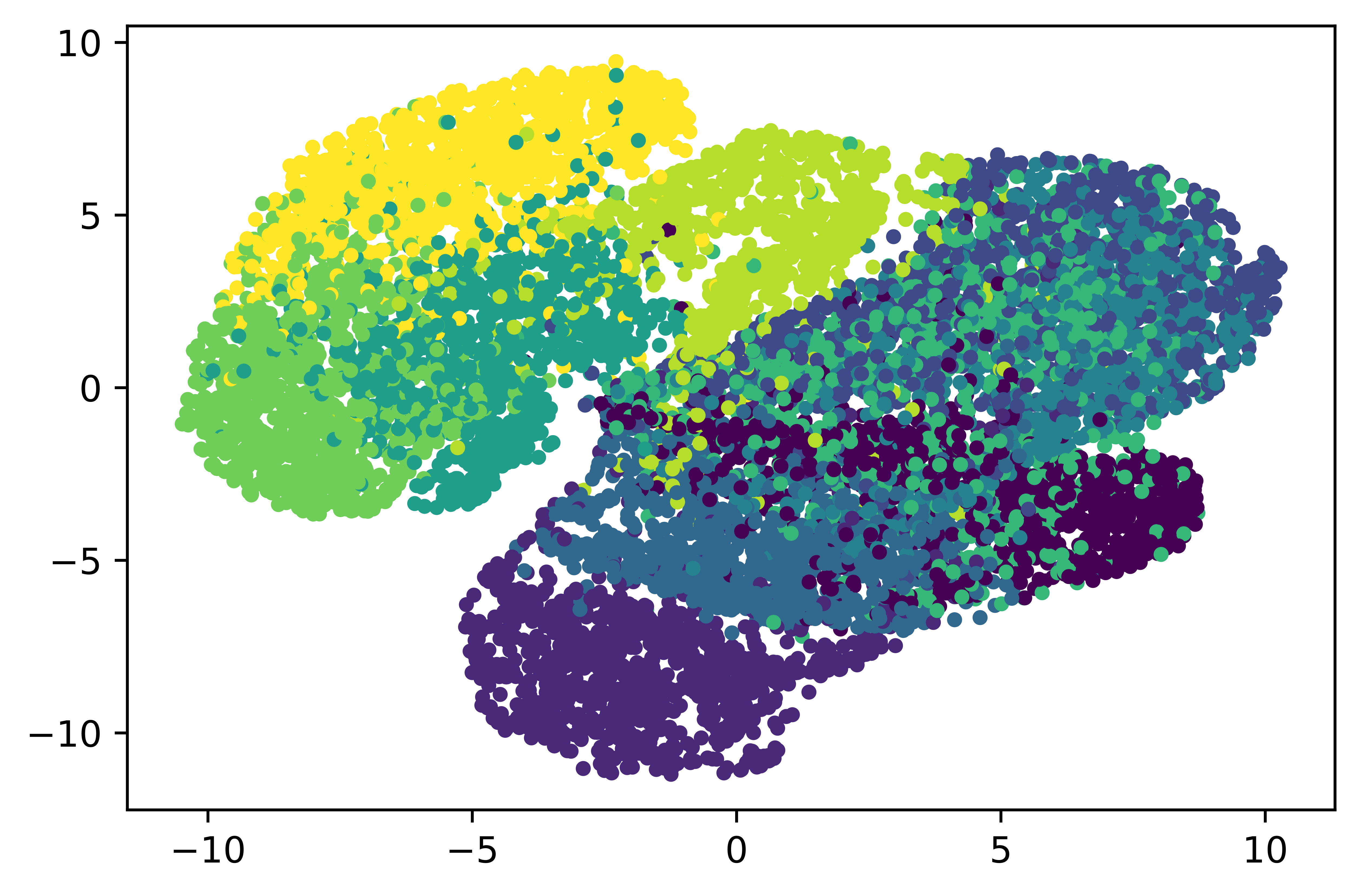}
SNE
\end{center}
\end{minipage}
\caption{Maps learned on the Fashion-MNIST dataset (in the 2D visualization space, colored by the labels)}
\label{fig:mapsmnist}
\end{figure}

We can see qualitatively that our maps have been learned properly for both methods in Fig.~\ref{fig:mapsmnist}, as they have grouped similar images into clusters (note that the labels used for coloring the points have not been used for training, but merely for this qualitative visualization). It is interesting to observe that the maps learned with SOM and SNE share similarities. The yellow and light green clusters are adjacent and well separated, and are also close to the two darker green clusters, that are more interrelated. (see left part of the maps in Fig.~\ref{fig:mapsmnist}). Except for the purple cluster that is clearly separated in both maps, the other clusters on the right hand sides of the maps learned seem very interrelated in both SOM and SNE maps. Such a qualitative evaluation shows that both maps are coherent with each other, which also supports our view that both approaches optimize a similar objective function.

\begin{figure}
\begin{minipage}{0.48\textwidth}
\begin{center}
\includegraphics[width=\textwidth]{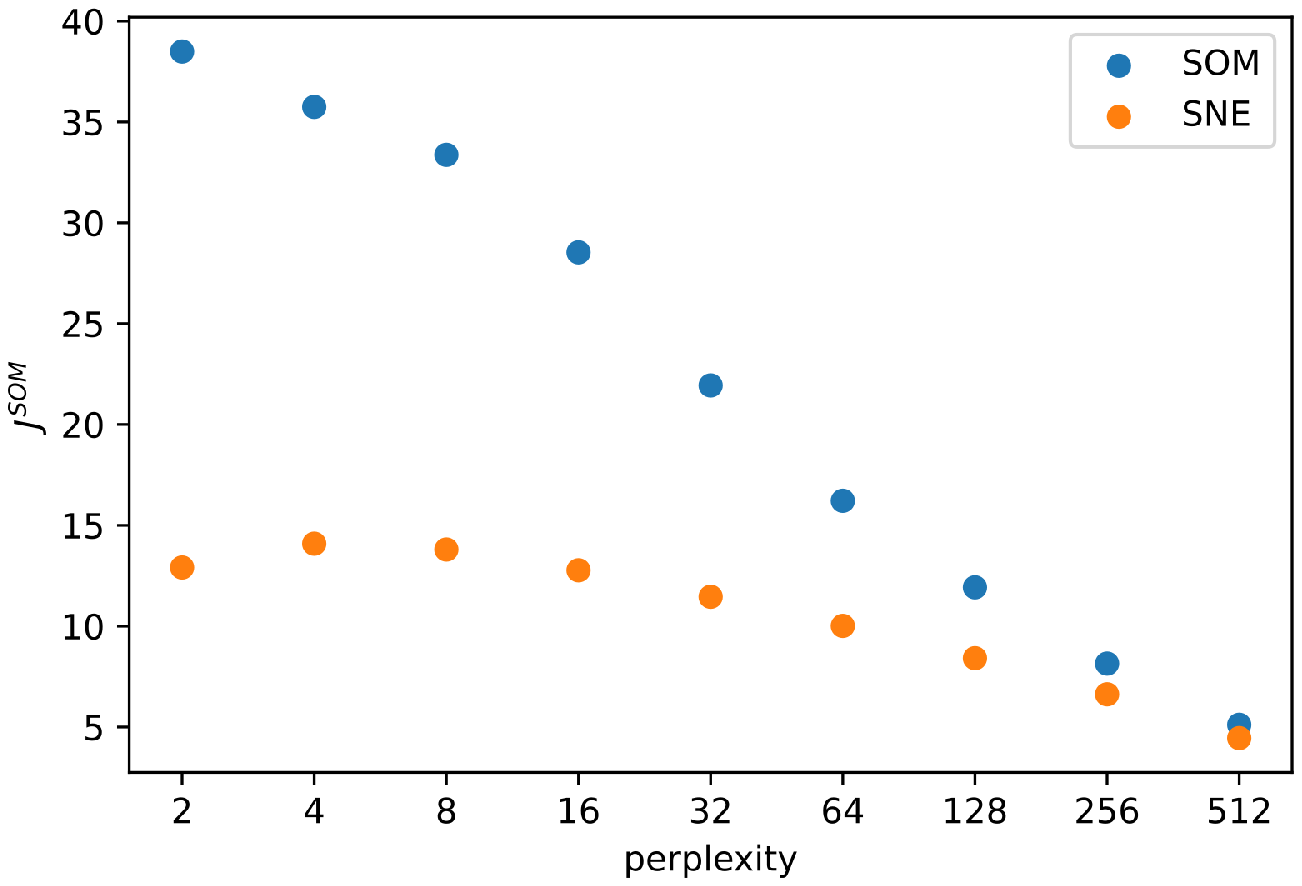}
\end{center}
\end{minipage}
\hspace{\stretch{1}}
\begin{minipage}{0.48\textwidth}
\begin{center}
\includegraphics[width=\textwidth]{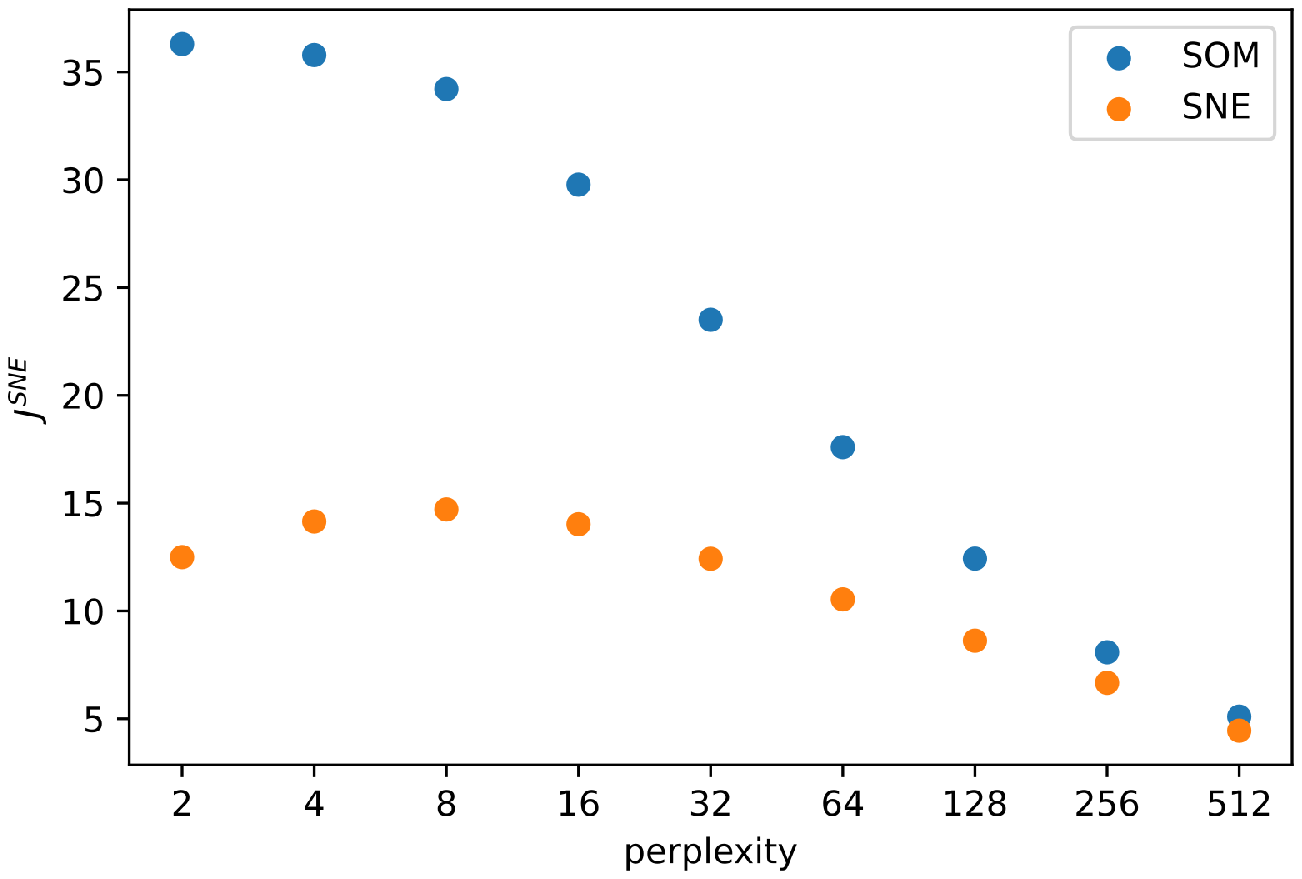}
\end{center}
\end{minipage}
\caption{Quantitative comparisons on the \textbf{Fashion-MNIST} dataset of the maps learned with SOM and SNE, for the two objectives we proposed (left: SOM objective, right: SNE objective)}
\label{fig:resultsmnist}
\end{figure}

We show the quantitative comparison in Fig.~\ref{fig:resultsmnist}, where we can see that the SNE algorithm outperforms the SOM algorithm for both objectives for all perplexities. This suggests that, even when using large maps, the assumption of the points lying on a grid in the visualization space might be too strong and unrealistic, which limits the quality of the SOM. Note that SNE outperforms SOM on both objectives considered, including the SOM objective. This is particularly impressive because SNE is not actually optimizing this very objective but a different one (with the forward KL divergence instead of the reverse KL divergence, see Table~\ref{table:comparison}).

\vspace{-2\baselineskip}

\newpage
\section{Conclusion}

In this work, we have proposed a unified view on the two widely used algorithms: Self-Organizing Maps (SOMs) and Stochastic Neighbor Embedding (SNE). We have shown that both approaches can be seen as the minimization of a very similar objective function attempting to organize the map and fit the data, the main difference being that the SOM algorithm fixes the 2D points (typically on a grid) and learns the neuron weights, while SNE fixes the neuron weights (typically, one for datapoint) and learns the 2D points. They can therefore be seen as dual algorithms, and can be compared quantitatively.

We proposed to compare them quantitatively on two datasets: a toy 2D dataset and the more realistic Fashion-MNIST dataset. In both experiments, we have observed that SNE significantly outperforms SOM for both objectives considered. We believe this is due to the SOM assumption that the 2D points are on a grid, which is very strong assumption that limits the quality of the map organization.

Most importantly, we hope that our work will inspire methods to combine the advantages of both approaches. SOMs are bio-inspired, have a fixed number of neurons, but are not biologically plausible due to the infinite neighborhood in the beginning. SNE, in its standard form, has as many neurons as the number of data points, which is not biologically plausible and might pose some problems in terms of complexity if it were to be applied online on infinitely growing datasets, but the perplexity is fixed throughout learning and no neighborhood parameter is required, which alleviates one of the fundamental drawbacks of SOM's algorithm: the need to tune and decay the neighborhood parameter carefully. Also, as we proposed in some of the experiments, our formulation of the SNE algorithm can be applied on a fixed number of neurons whose weights have been learned previously (e.g., using the kMeans algorithm).

Methods jointly learning both the positions in 2D and the neuron weights could definitely be worth considering, and open an interesting avenue for future work. 

%
%
%
\bibliographystyle{splncs04}
\bibliography{sample}
%




\end{document}